\documentclass[runningheads]{llncs}
\usepackage{graphicx}
\usepackage{amsmath,amssymb} 
\usepackage{color}
\usepackage{colortbl}
\usepackage{multirow}
\usepackage{multicol}
\usepackage{arydshln}
\usepackage{adjustbox}
\usepackage{enumerate}
\usepackage{multicol}
\usepackage{placeins}
\usepackage{blindtext}
\usepackage{booktabs}

\definecolor{ForestGreen}{rgb}{0.13, 0.55, 0.13}

\definecolor{gray}{gray}{0.7}
\definecolor{red}{rgb}{0.9, 0.3, 0.3}
\definecolor{green}{rgb}{0.3, 0.9, 0.3}

\usepackage[width=122mm,left=12mm,paperwidth=146mm,height=193mm,top=12mm,paperheight=217mm]{geometry}

\begin{document}
\pagestyle{headings}
\mainmatter

\title{Revisiting Visual Question Answering Baselines} 

\titlerunning{Revisiting Visual Question Answering Baselines}

\authorrunning{Jabri, Joulin, and van der Maaten}

\author{Allan Jabri, Armand Joulin, and Laurens van der Maaten}
\institute{Facebook AI Research\\
	\email{\{ajabri,ajoulin,lvdmaaten\}@fb.com}
}

\maketitle

\begin{abstract} Visual question answering (VQA) is an interesting learning
setting for evaluating the abilities and shortcomings of current systems for
image understanding. Many of the recently proposed VQA systems include attention or memory mechanisms 
designed to support ``reasoning''. For multiple-choice VQA, nearly all of
these systems train a multi-class classifier on image and question features
to predict an answer. This paper questions the value of these common practices and develops
a simple alternative model based on binary classification. Instead of treating
answers as competing choices, our model receives the answer as input and predicts whether or not an image-question-answer triplet is correct. We
evaluate our model on the Visual7W Telling and the VQA Real Multiple Choice
tasks, and find that even simple versions of our model perform competitively.
Our best model achieves state-of-the-art performance on the
Visual7W Telling task and compares surprisingly well with the most complex
systems proposed for the VQA Real Multiple Choice task. We
explore variants of the model and study its transferability between both datasets. We also present an error analysis of our model that suggests a key problem of current VQA systems lies in
the lack of visual grounding of concepts that occur in the
questions and answers. Overall, our results suggest that the performance of 
current VQA systems is not significantly better than that of systems designed to exploit dataset biases.
\end{abstract}

\section{Introduction}
\label{Introduction}
Recent advances in computer vision have brought us close to the point
where traditional object-recognition benchmarks such as
Imagenet are considered to be~``solved''~\cite{szegedy15,he16}. These
advances, however, also prompt the question how we can move from
object recognition to \emph{visual understanding}; that is, how we
can extend today's recognition systems that provide us with
``words'' describing an image or an image region to systems that
can produce a deeper semantic representation of the image content.
Because benchmarks have traditionally been a key driver for progress
in computer vision, several recent studies have proposed
methodologies to assess our ability to develop such representations.
These proposals include modeling relations between objects
\cite{krishna16}, visual Turing tests~\cite{geman15}, and visual
question answering~\cite{antol15,ren15,yu15,zhu15}.


\begin{figure}[h!]
	\scriptsize
	\centering
	\begin{tabular}{m{0.22\textwidth}cm{0.22\textwidth}cm{0.22\textwidth}cm{0.22\textwidth}}
	\includegraphics[width=0.22\textwidth,height=.18\textwidth]{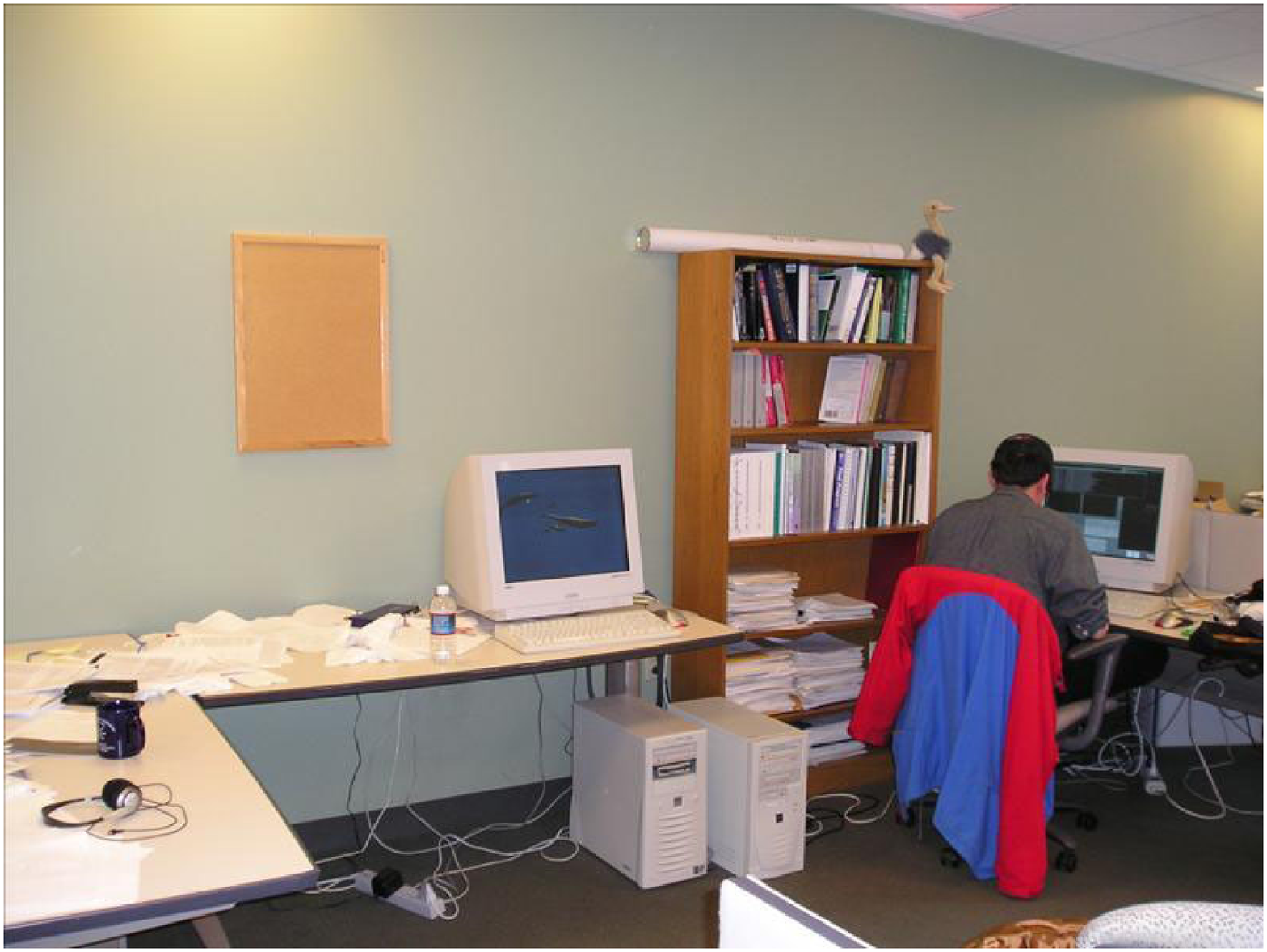} 
	&~~~&
	\includegraphics[width=0.22\textwidth,height=.18\textwidth]{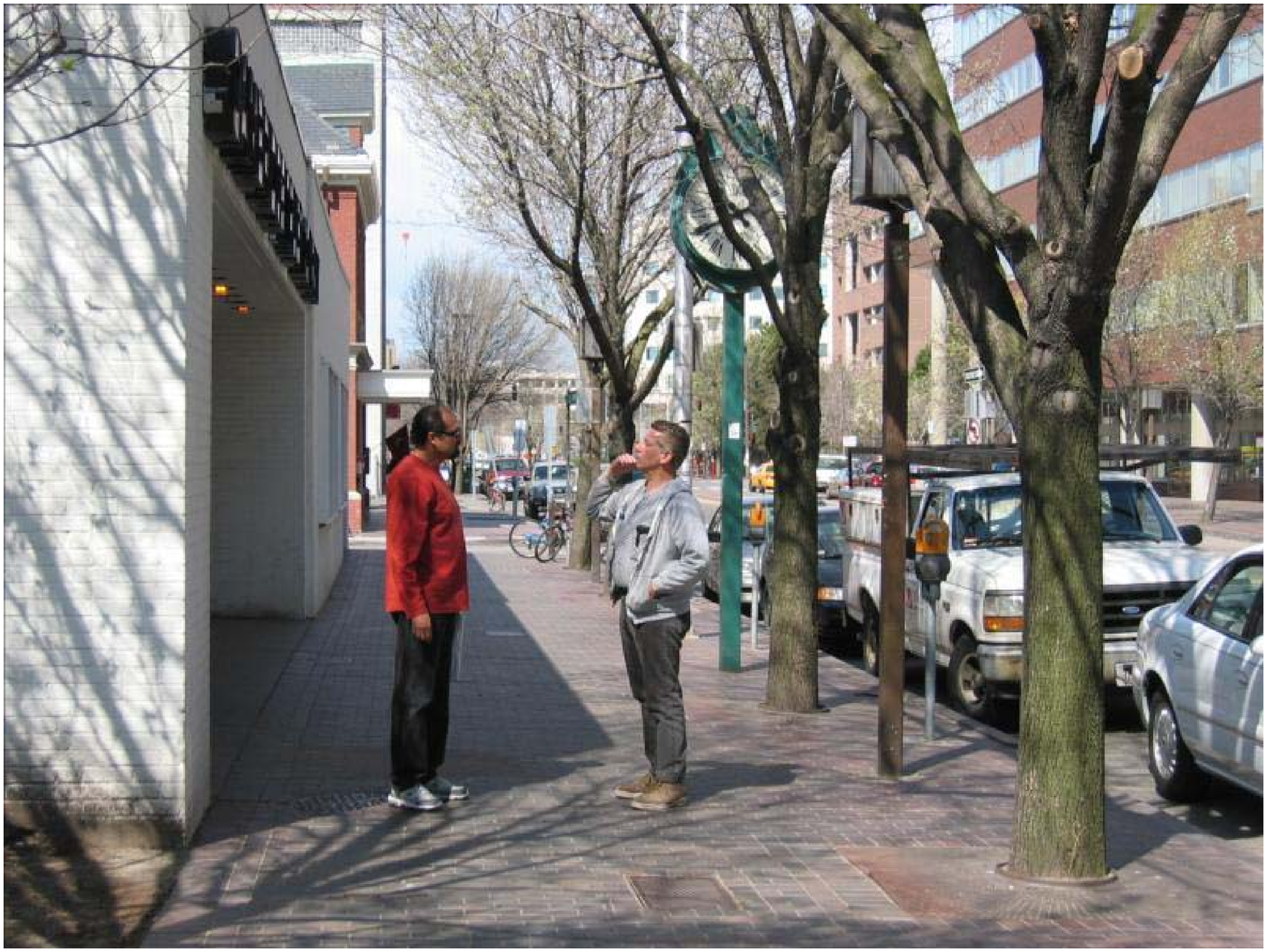} 
	&~~~&
	\includegraphics[width=0.22\textwidth,height=.18\textwidth]{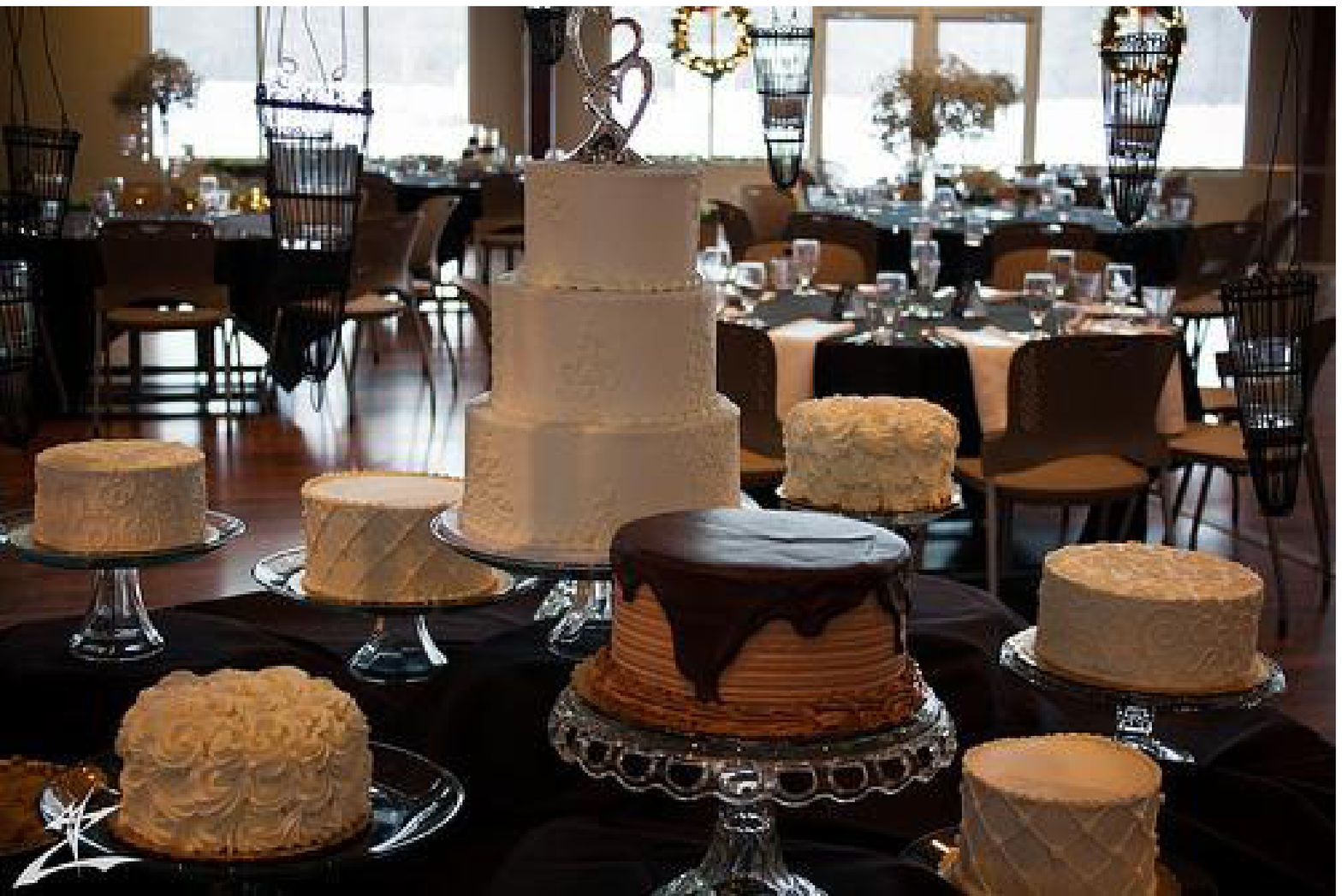} 
	&~~~&
	\includegraphics[width=0.22\textwidth,height=.18\textwidth]{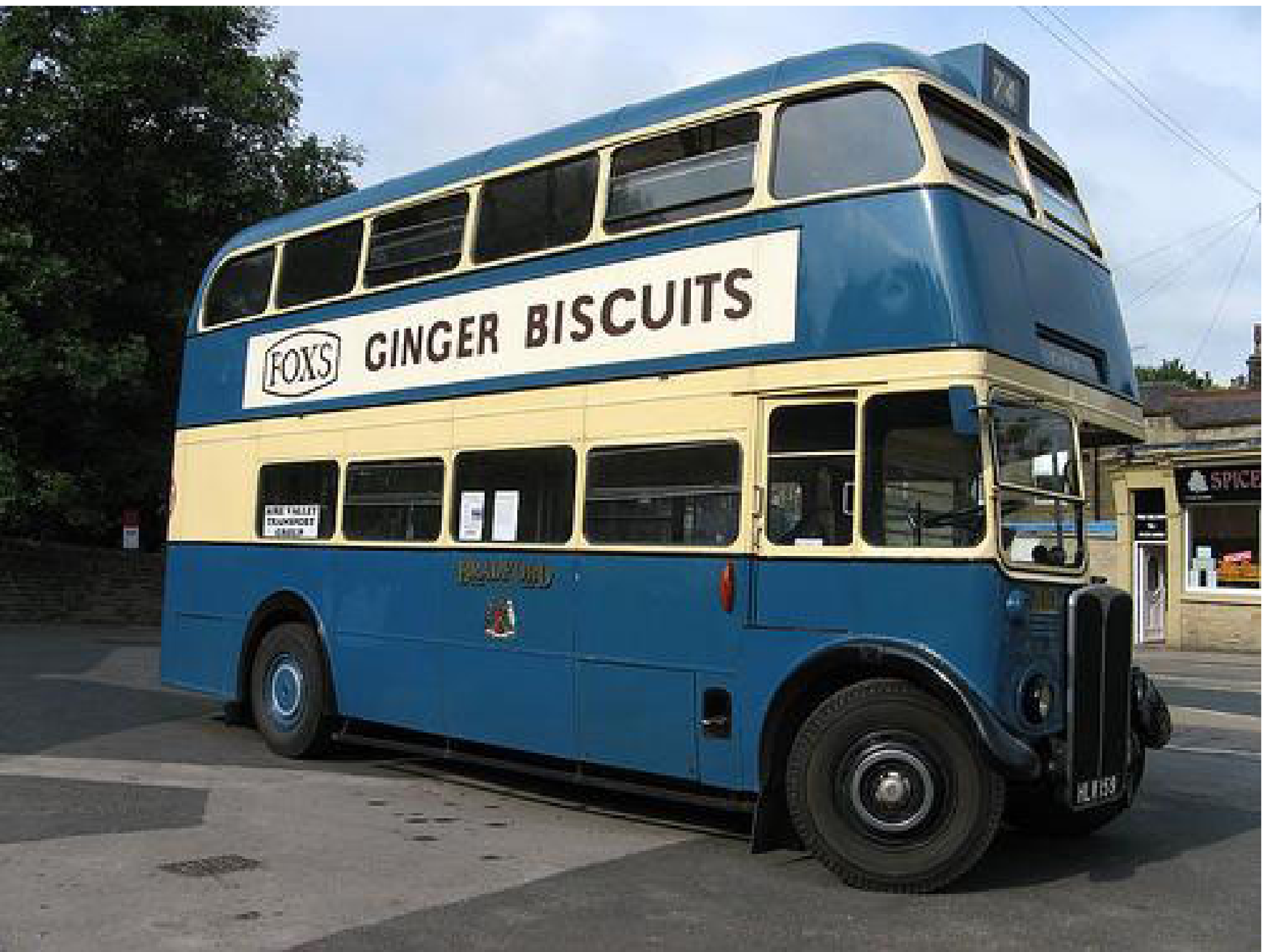} 
	\\
	What color is the jacket? && How many cars are parked?
	&&
	What event is this? && When is this scene taking place?
	\\
	{\color{ForestGreen}-Red and blue.} && {\color{ForestGreen}-Four.} 
	&&
	{\color{ForestGreen}-A wedding.} && {\color{ForestGreen}-Day time.} 
	\\
	-Yellow. && -Three.
	&&
	-Graduation. && -Night time.
	\\
	-Black. && -Five.
	&&
	-A funeral. &&  -Evening.
	\\
	-Orange. && -Six.
	&&
	-A picnic. && -Morning.
	\end{tabular}
	\caption{Four images with associated questions and answers from the Visual7W dataset. Correct answers are typeset in green.}\label{fig:vqa_examples}
    \kern-20pt
\end{figure}

The task of Visual Question Answering (VQA) is to answer questions---posed in
natural language---about an image by providing an answer in the form of short~text.  This answer can either be selected from multiple pre-specified choices
or be generated by the system.
As can be seen from the examples in Figure~\ref{fig:vqa_examples},
VQA combines computer vision with natural
language processing and reasoning.

VQA seems to be a natural playground to develop approaches able to perform basic
``reasoning'' about an image.  Recently, many studies have explored this
direction by adding simple memory or attention-based components to VQA systems.  While in theory, these
approaches have the potential to perform simple reasoning, it is not
clear if they do actually reason, or if they do so in a human-comprehensible way. For
example,
Das et al.~\cite{das2016human} recently reported that ``machine-generated
attention maps are either negatively correlated with human attention or
have positive correlation worse than task-independent  saliency''.
In this work, we also question the significance of the performance
obtained by current ``reasoning''-based systems. In particular, this study sets out to
answer a simple question: are these systems better than baselines
designed to solely capture the dataset bias of standard VQA datasets?
We limit the scope of our study to multiple-choice tasks, as this allows us to perform a
more controlled study that is not hampered by the tricky nuances of
evaluating generated text~\cite{koehn2004statistical,callison2006re}. 

We perform experimental evaluations on the Visual7W dataset~\cite{zhu15} and
the VQA dataset~\cite{antol15} to evaluate the quality of our baseline models.
We: (1)~study and model the bias in the Visual7W Telling and VQA Multiple Choice
datasets, (2)~measure the effect of using visual features from different CNN
architectures, (3)~explore the use of a LSTM as the system's language model,
  and (4)~study transferability of our model between datasets. 

Our best model outperforms the current state-of-the-art on the
Visual7W telling task with a performance of $67.1\%$, and competes surprisingly well with the most complex
systems proposed for the VQA dataset. Furthermore, our models perform competitively even with missing information 
(that is, missing images, missing questions, or both). Taken together, our results suggests that the performance of 
current VQA systems is not significantly better than that of systems designed to exploit dataset biases.


\section{Related work}
\label{Related work}
The recent surge of studies on visual question answering has been fueled by the
release of several visual question-answering datasets, most prominently, the
VQA dataset~\cite{antol15}, the DAQUAR dataset ~\cite{daquar}, the Visual Madlibs Q\&A dataset~\cite{yu15}, the
Toronto COCO-QA dataset~\cite{ren15}, and the Visual7W dataset~\cite{zhu15}. 
Most of these datasets were developed by annotating subsets of the
COCO dataset~\cite{lin14}. Geman~\emph{et~al.}~\cite{geman15} proposed a visual
Turing test in which the questions are automatically generated and require no
natural language processing. Current approaches to visual question answering
can be subdivided into~``generation'' and~``classification'' models:

\kern2mm
\noindent\textbf{Generation models.}
Malinowski~\emph{et~al.}~\cite{malinowski15} train a LSTM model to generate
the answer after receiving the image features~(obtained from a convolutional
    network) and the question as input. Wu~\emph{et~al.}~\cite{wu16} extend a
LSTM generation model to use external knowledge that is obtained from
DBpedia~\cite{auer07}.  Gao~\emph{et~al.}~\cite{gao15} study a similar model
but decouple the LSTMs used for encoding and decoding. Whilst generation models
are appealing because they can generate arbitrary answers~(also answers that
    were not observed during training), in practice, it is very difficult to
jointly learn the encoding and decoding models from the question-answering
datasets of limited size. In addition, the evaluation of the quality of the
generated text is complicated in practice~\cite{koehn2004statistical,callison2006re}. 

\kern2mm
\noindent\textbf{Classification models.} Zhou~\emph{et~al.}~\cite{zhou15} study
an architecture in which image features are produced by a convolutional
network, question features are produced by averaging word embeddings over all
words in the question, and a multi-class logistic regressor is trained on the
concatenated features; the top unique answers are
treated as outputs of the classification model. Similar approaches are
also studied by Antol~\emph{et~al.}~\cite{antol15} and
Ren~\emph{et~al.}~\cite{ren15}, though they use a LSTM to encode the question text instead of an average over word embeddings. Zhu~\emph{et~al.}~\cite{zhu15}
present a similar method but extend the LSTM encoder to include an attention mechanism for jointly encoding the question with information from the image. Ma~\emph{et~al.}~\cite{ma15} replace the LSTM
encoder by a
one-dimensional convolutional network that combines the word embeddings into a
question embedding. Andreas~\emph{et~al.}~\cite{andreas15} use a similar model
but perform the image processing using a compositional network whose structure
is dynamically determined at run-time based on a parse of the question.
Fukui~\emph{et~al.}~\cite{mcb} propose the use of~``bilinear pooling'' for
combining multi-modal
information. Lu~\emph{et~al.}~\cite{hiatt16} jointly learn a hierarchical attention
mechanism based on parses of the question and the image which they call ``question-image co-attention''. 

\kern2mm
\noindent Our study is similar to a recent study by
Shih~\emph{et~al.}~\cite{wtl16}, which also considers models that treat the answer as an input variable and predicts whether or not an image-question-answer triplet is correct. However, their study develops a substantially more complex pipeline involving image-region selection while achieving worse performance.

\begin{figure}[t]
	\centering
	\includegraphics[width=.85\textwidth]{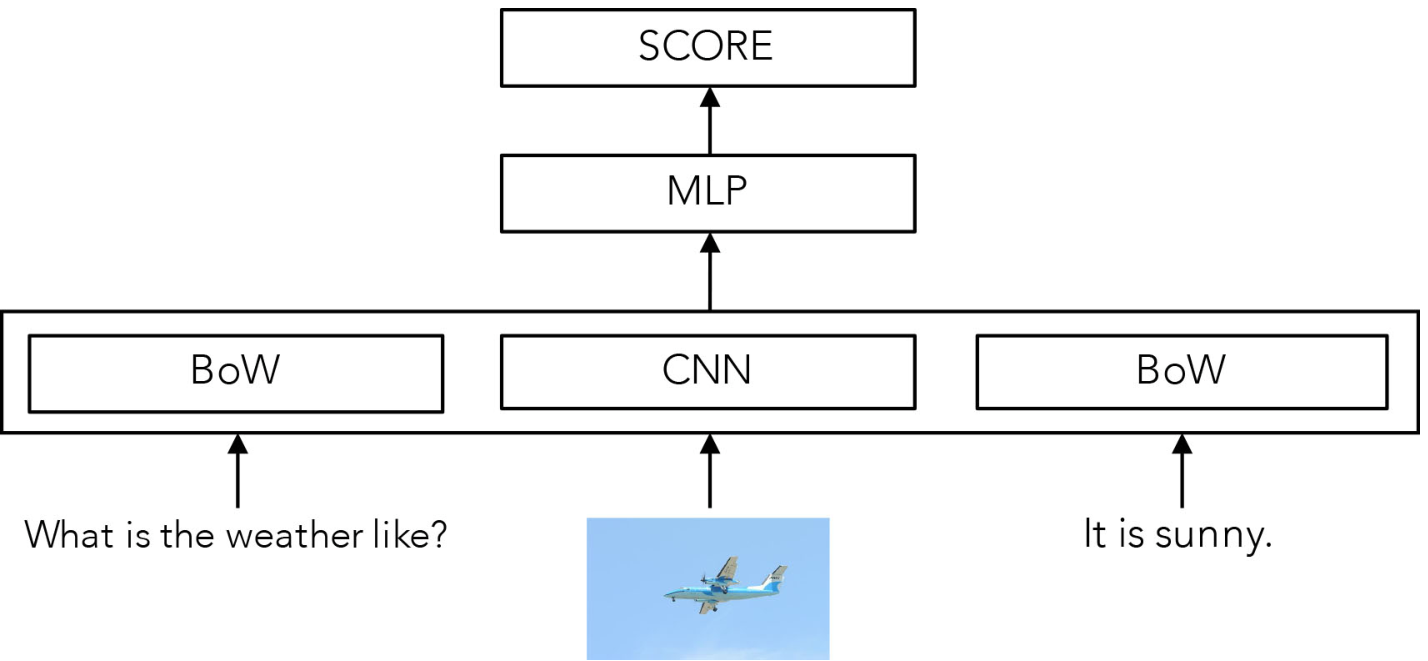}
	\caption{Overview of our system for visual question answering. See text for
		details.}\label{fig:system_overview}
\kern-10pt
\end{figure}
\FloatBarrier

\section{System Overview}
\label{System Overview}
Figure~\ref{fig:system_overview} provides an overview of the architecture of
our visual question answering system. The system takes an image-question-answer
feature triplet as input. Unless otherwise stated (that is, in the LSTM
    experiment of Section~\ref{Experiments}), both the questions and the
answers are represented by averaging word2vec embeddings over all words in the
question or answer, respectively. The images are represented using features
computed by a pre-trained convolutional network. Unless otherwise stated, we
use the penultimate layer of Resnet-101~\cite{he16}.  The word2vec embeddings
are~$300$-dimensional and the image features are~$2,048$-dimensional. The three
feature sets are concatenated and used to train a classification model that
predicts whether or not the image-question-answer triplet is correct.

The classification models we consider are logistic regressors and multilayer
perceptrons~(MLP) trained on the concatenated features, and bilinear models
that are trained on the answer features and a concatenation of the image and
question features. The MLP has~$8,192$ hidden units unless otherwise specified. We use dropout~\cite{dropout} after the first layer.
We denote the image, question, and answer features 
by~$\mathbf{x}_i$,~$\mathbf{x}_q$, and~$\mathbf{x}_a$, respectively. 
Denoting the sigmoid function~$\sigma(x) = 1 / (1 + \exp(-x))$ 
and the concatenation operator~$\mathbf{x}_{iq} = \mathbf{x}_i \oplus \mathbf{x}_q$, we define the models as follows:

\kern5pt
\hspace{24mm} $\begin{aligned}[t]
\textrm{\textbf{Linear:}} \indent &  y = \sigma(\mathbf{Wx}_{iqa} + b) \\
\textrm{\textbf{Bilinear:}} \indent & y =  \sigma(\mathbf{x}^\top_{iq}\mathbf{Wx}_a + b) \\
\textrm{\textbf{MLP:}} \indent & y =  \sigma( \mathbf{W}_2 \textrm{ max}(0, \mathbf{W}_1 \mathbf{x}_{iqa}) + b).\\
\end{aligned}
$

\kern2mm
The parameters of the classifier are learned by minimizing the binary logistic loss of predicting whether or not an image-question-answer triplet is correct using stochastic gradient
descent. During training we sampled two negative examples from the multiple choices for each positive example, for a maximum of 300 epochs. The convolutional networks were
pre-trained on the Imagenet dataset, following~\cite{samresnet}, and were not further finetuned.
We used pre-trained word2vec~\cite{mikolov13} embeddings, which we did not finetune on VQA data either.

%

%

\section{Experiments}
\label{Experiments}

\vspace{-2mm}

We perform experiments on the following two datasets: 

\kern2mm
\noindent\textbf{Visual7W Telling~\cite{zhu15}.} The dataset includes~$69,817$ training
questions,~$28,020$ validation questions, and~$42,031$
test questions. Each question has four answer choices. The negative choices are
human-generated on a per-question basis. The performance is measured
by the percentage of correctly answered questions.

\kern2mm
\noindent\textbf{VQA Real Multiple Choice~\cite{antol15}.} The dataset includes~$248,349$
questions for training,~$121,512$ for validation, and~$244,302$ for testing. 
Each question has~$18$ answer choices. The negative choices are
randomly sampled from a predefined set of answers.
Performance is measured following the metric proposed by~\cite{antol15}.

\begin{table}[t]
	\caption{Comparison of our models with the state-of-the-art for the Visual7W telling task \cite{zhu15}. Human accuracy on the task is $96.0\%$. Higher values are better.}
	\label{table:v7w}
	\begin{center}
		\begin{small}
			\def\arraystretch{1.0}
			\setlength\tabcolsep{2.5pt}
			\begin{tabular}{lccccccc}
				\toprule
				\textbf{Method} & ~\textbf{What}~ & ~\textbf{Where}~ & ~\textbf{When}~ & ~\textbf{Who}~ & ~\textbf{Why}~ & ~\textbf{How}~ & \textbf{Overall} \\
				\midrule
				LSTM (Q, I) \cite{malinowski15}~ & $48.9$ & $54.4$ & $71.3$ & $58.1$ & $51.3$ & ${50.3}$ & $52.1$ \\
				LSTM-Att \cite{zhu15} & $51.5$ & $57.0$ & $75.0$ & $59.5$ & $55.5$ & $49.8$ & $55.6$ \\
				MCB + Att \cite{mcb} & $60.3$ & $70.4$ & $79.5$ & $69.2$ & $58.2$ & 51.1 & $62.2$ \\
				\midrule
				Bilinear (A, Q, I) & $60.4$ & $72.3$ & $78.0$ & $71.6$ & $63.0$ & $54.8$ & $63.6$ \\
				MLP (A)  & $47.3$ & $58.2$ & $74.3$ & $63.6$ & $57.1$ & $49.6$ & $52.9$ \\
				MLP (A, Q)  & $54.9$ & $60.0$ & $76.8$ & $66.0$ & $64.5$ & $54.9$ & $58.5$ \\
				MLP (A, I) & $60.8$ & $74.9$ & $81.9$ & $70.3$ & $64.4$ & $51.2$ & $63.8$ \\
				MLP (A, Q, I)~~~ & $\textbf{64.5}$ & $\textbf{75.9}$ & $\textbf{82.1}$ & $\textbf{72.9}$ & $\textbf{68.0}$ & $\textbf{56.4}$ & $\textbf{67.1}$  \\
				\bottomrule
			\end{tabular}
		\end{small}
	\end{center}
	\kern-10pt
\end{table}

\begin{table}[t]
	\caption{Comparison of our models with the state-of-the-art single models for the VQA Real Multiple Choice task \cite{antol15}. Results are reported on the test2015-standard split. Human accuracy on the task is $83.3\%$. * refers to results on test2015-dev.}
	\label{table:vqa}
	\begin{center}
		\begin{small}
			\def\arraystretch{1.0}
			\setlength\tabcolsep{2.5pt}
			\begin{tabular}{lcccc}
				\toprule
				\textbf{Method} & ~\textbf{Yes/No}~ & ~\textbf{Number}~ & ~\textbf{Other}~ & ~\textbf{All}~ \\
				\midrule
				Two-Layer LSTM \cite{antol15} & $80.6$ & $37.7$ & $53.6$ & $63.1$ \\
				Region selection~\cite{wtl16} & 77.2 & 33.5 & 56.1 & 62.4\\
				Question-Image Co-Attention \cite{hiatt16}~~~ & $80.0$ & \textbf{39.5} & $59.9$ & 66.1 \\
				MCB \cite{mcb}* & -- & -- & -- & 65.4 \\
				MCB + Att + GloVe + Genome \cite{mcb}* & -- & -- & -- & \textbf{69.9} \\
				Multi-modal Residual Network \cite{naver} & -- & -- & -- & 69.3 \\
				\midrule
				MLP (A, Q, I) & \textbf{80.8} & $17.6$ & \textbf{62.0} & $65.2$  \\
				\bottomrule
			\end{tabular}
		\end{small}
	\end{center}
	\kern-10pt
\end{table}

\kern-5pt
\subsection{Comparison with State-of-the-Art}

\vspace{-2mm}

We first compare the MLP variant of our model with the state-of-the-art. 
Table~\ref{table:v7w} shows the results of this comparison on Visual7W, using three variants of our baseline
with different inputs:~(1) answer and question~(A+Q);~(2) answer and image~(A+I);~(3)
and all three inputs~(A+Q+I). 
The model achieves state-of-the-art performance when it has
access to all the information. 
Interestingly, as shown by the results with the A+Q variant of our model,
  simply exploiting the most frequent question-answer pairs obtains competitive
  performance. Surprisingly, even a variant of our model that is trained
  \emph{on just the answers} already achieves a performance of~$52.9\%$, simply
  by learning biases in the answer distribution.

In Table~\ref{table:vqa}, we also compare our models with the published state-of-the-art on
the VQA dataset. Despite its simplicity, our baseline achieves comparable performance 
with state-of-the-art models. We note that recent state-of-the-art work~\cite{mcb}
used an ensemble of~$7$ models
trained on additional data~(the Visual Genome dataset~\cite{krishna16}), 
performing $5\%$ better than our model whilst being
substantially more complex.

\kern-5pt
\subsection{Additional Experiments}
In the following, we present the results of additional experiments to
understand why our model performs relatively well, and when it fails. All
evaluations are conducted on the Visual7W Telling dataset unless stated
otherwise.
 


\begin{table}
	\kern-10pt
	\begin{minipage}{0.45\linewidth}
	\begin{tabular}{llcc}
	\toprule
	\textbf{Dataset} &  \textbf{Model} & ~~\textbf{Softmax}~~ & \textbf{Binary}\\
	\midrule
	\multirow{3}{*}{Visual7W~~~} &  Linear     & $42.6$ & $44.7$ \\ 
	&  Bilinear~~~~& --     & $63.6$ \\ 
	&  MLP        & $52.2$ & $\textbf{67.1}$ \\ 
	\midrule
	VQA & MLP & 61.1 & \textbf{64.9}\\
	\bottomrule
	  \end{tabular}
	\end{minipage}
	\hspace{14mm}
	\begin{minipage}{0.4\linewidth}
	\caption{Accuracy of models using either a \emph{softmax} or a \emph{binary} loss. 
	  Results are presented for different models using answer, question and image. On VQA,  we use the test2015-dev split. Higher values are better.}
	\end{minipage}
	\label{table:binary_results}
	\kern-20pt
\end{table}

\begin{table}
	  \kern-30pt
	\caption{The five most similar answers in the Visual7W dataset for three answers appearing in that dataset (in terms of cosine similarity between their feature vectors).}
	\label{table:sim}
	\begin{center}
		\begin{small}
			\begin{tabular}{lll}
      \toprule
        During the daytime. & On the bus stop bench.   & On a tree branch. \\
        \midrule
        During daytime. & Bus bench. & On the tree branch. \\
        Outside, during the daytime.~~~~ & In front of the bus stop. & The tree branch. \\
        Inside, during the daytime. & The bus stop. & Tree branch. \\
        In the daytime. & At the bus stop. & A tree branch.\\
        In the Daytime. & The sign on the bus stop.~~~~ & Tree branches.\\
        \bottomrule
			\end{tabular}
		\end{small}
	\end{center}
  \kern-20pt
\end{table}

\noindent\textbf{Does it help to consider the answer as an input?}
In Table~\ref{table:binary_results}, we present the results of experiments in
  which we compare the performance of our~(binary) baseline model with variants
  of the model that predict softmax probabilities over a discrete set of
  the~$5,000$ most common answers, as is commonly done in most prior
  studies, for instance,~\cite{zhou15}.

The results in the table show a substantial advantage
of representing answers as inputs instead of outputs for the Visual7W Telling task
and the VQA Real Multiple Choice task. 
Taking the answer as an input allows the system to model the similarity between different
answers.  For example, the answers~``two people'' and~``two persons'' are
modeled by disjoint parameters in a softmax model, whereas the binary model
will assign similar scores to these answers because they have similar
bag-of-words word2vec representations.
 
To illustrate this, Table~\ref{table:sim} shows examples of the similarities
captured by the BoW representation. For a given answer, the table shows
the five most similar answers in the dataset based on cosine similarity between
the feature vectors. The binary model can readily exploit these similarities, 
whereas a softmax model has to learn them from the (relatively small) Visual7W training set.
  
Interestingly, the gap between the binary and softmax models is
smaller on the VQA datasets. This result may be explained by the
way the incorrect-answer choices were produced in both datasets: the choices are
human-generated for each question in the Visual7W dataset, whereas in the VQA dataset,
the choices are randomly chosen from a predefined set that includes irrelevant correct answers.

\begin{table}[t]
\caption{Accuracy on the Visual7W Telling task using visual features produced by five different convolutional networks. Higher values are better.}
\label{table:convnet_results}
	\begin{center}
	\begin{small}
	\def\arraystretch{1.1}
	\setlength\tabcolsep{2.5pt}
	\setlength\tabcolsep{2.5pt}
	\begin{tabular}{lccccc}
	\toprule
	\textbf{Model} & \textbf{AlexNet}~ & ~\textbf{GoogLeNet}~ & ~\textbf{ResNet-34}~ & ~\textbf{ResNet-50}~ & ~\textbf{ResNet-101}\\
	\midrule
	(dim.) & (4,096) & (1,792) & (512) & (2,048) & (2,048)\\
	Bilinear~~~ & $56.3$ & $58.5$ & $60.1$ & $62.4$ & $63.6$\\
	MLP  & $63.5$ & $64.2$ & $65.9$ & $66.3$ & $\mathbf{67.1}$\\
	\bottomrule
	\end{tabular}
	\end{small}
	\end{center}
\kern-20pt
\end{table}

\kern3mm
\noindent\textbf{What is the influence of convolutional network architectures?}
Nearly all prior work on VQA uses features extracted using a convolutional
network that is pre-trained on Imagenet to represent the image in an
image-question pair.  Table~\ref{table:convnet_results} shows to what extent
the quality of these features influences the VQA performance by comparing
five different convolutional network architectures:
AlexNet~\cite{krizhevsky12}, GoogLeNet~\cite{szegedy15}, and residual
networks with three different depths~\cite{he16}.  While the performance on
Imagenet is correlated with performance in visual question answering, the
results show this correlation is quite weak: a reduction in the Imagenet
top-5 error of~$18\%$ corresponds to an improvement of only~$3\%$ in
question-answering performance.  This result suggests that the performance on
VQA tasks is limited by either the fact that some of the visual concepts in
the questions do not appear in Imagenet, or by the fact that the
convolutional networks are only trained to recognize object presence and not
to predict higher-level information about the visual content of the images.

\begin{table}
	\begin{minipage}{0.3\linewidth}
	\kern-20pt
	\def\arraystretch{1.1}
	\setlength\tabcolsep{2.5pt}
		\begin{tabular}{lcc}
		\toprule
		\textbf{Model} & ~~\textbf{BoW}~~ & \textbf{LSTM}\\
		\midrule
		Bilinear~~~ & $52.6$ & $54.3$ \\
		MLP & $\mathbf{58.5}$ & $52.9$ \\
		\bottomrule
		\end{tabular}
	\end{minipage}
	\hspace{10mm}
	\begin{minipage}{0.6\linewidth}
		\kern-10pt
		\caption{ Accuracy on Visual7W Telling dataset of a bag-of-words (\emph{BoW}) and a \emph{LSTM} model. We did not use image features to
			isolate the difference between language models. Higher values are better.}
		\label{table:lstm_results}
	\end{minipage}
\kern-10pt
\end{table}
\kern-20pt

\noindent\textbf{Do recurrent networks improve over bag of words?}
Our baseline uses a simple bag-of-words~(BoW) model to represent the questions
and answers. Recurrent networks~(in particular, LSTMs~\cite{hochreiter97})
are a popular alternative for BoW models. We perform an experiment in which we
replace our BoW representations by a LSTM model. The LSTM was trained on the
Visual7W Telling training set, using a concatenation of one-hot encodings and
pre-trained word2vec embeddings as input for each word in the question.

For the final representation, we observed little difference between using the average
over time of the hidden states versus using only the last hidden
state. Here, we report the results using the last-state representation.

Table~\ref{table:lstm_results} presents the results of our experiment comparing
BoW and LSTM representations. To isolate the difference between the language
models, we did not use images features as input in this experiment. The results
show that despite their greater representation power, LSTMs actually do not
outperform BoW representations on the Visual7W Telling task, presumably, because the
dataset is quite small and the LSTM overfits easily. This may also explain why
attentional LSTM models~\cite{zhu15} perform poorly on the Visual7W dataset.

\begin{table}[t]
	\caption{Accuracy on Visual7W of models (1) trained from \emph{scratch}, (2) \emph{transfer}ed from the VQA dataset, and (3) \emph{finetune}d after transferring. Higher values are better.}
	\label{table:trans_results}
	\kern-10pt
	\begin{center}
		\begin{small}
			\def\arraystretch{1.1}
			\setlength\tabcolsep{2.5pt}
			\begin{tabular}{llccccccc}
				\toprule

				\textbf{Model} &\textbf{Method}~~ & ~\textbf{What}~ & ~\textbf{Where}~ & ~\textbf{When}~ & ~\textbf{Who}~ & ~\textbf{Why}~ & ~\textbf{How}~ & \textbf{Overall} \\ 
        \midrule
				
				\multirow{2}{5em}{MLP \\(A+Q)} &Scratch & $54.9$ & $60.0$ & $76.8$ & $66.0$ & $64.5$ & $54.9$ & $58.5$  \\
				
				&Transfer  & $44.7$ & $38.9$ & $32.9$ & $49.6$ & $45.0$ & $27.3$ & $41.1$   \\
        \midrule
				
				\multirow{2}{5em}{MLP \\(A+I)} &Scratch & $60.8$ & $74.9$ & $81.9$ & $70.3$ & $64.4$ & $51.2$ & $63.8$  \\
				
				&Transfer  &  $28.4$  & $26.6$  & $44.1$  & $37.0$  & $31.7$  & $25.2$  & $29.4$    \\
        \midrule	
				
				\multirow{3}{5em}{MLP \\(A+Q+I)}~ &Scratch & $64.5$ & $75.9$ & $82.1$ & $72.9$ & $68.0$ & $56.4$ & $67.1$  \\
				
				&Transfer  & $58.7$ & $61.7$ & $41.7$ & $60.2$ & $53.2$ & $29.1$ & $53.8$  \\
				
				&Finetune & $\textbf{66.4}$ & $\textbf{77.1}$ & $\textbf{83.2}$ & $\textbf{73.9}$ & $\textbf{70.7}$ & $\textbf{56.7}$ & $\textbf{68.5}$ \\
        \bottomrule
			
			\end{tabular}
		\end{small}
	\end{center}
	\kern-30pt
\end{table}

\kern2mm
\noindent\textbf{Can we transfer knowledge from VQA to Visual7W?}
An advantage of the presented model is that it can readily be transfered between datasets:
it does not suffer from out-of-vocabulary problems nor does it require
the set of answers to be known in advance.
Table~\ref{table:trans_results} shows the results of a transfer-learning
experiment in which we train our model on the VQA dataset, and use it to answer
questions in the Visual7W dataset. We used three different variants of our model, and experimented with
    three different input sets. The table presents three sets of results:~(1)
baseline results in which we trained on Visual7W from~\emph{scratch},~(2)~\emph{transfer} 
results in which we train on VQA but test on Visual7W, and~(3)
results in which we train on VQA,~\emph{finetune} on Visual7W, and then test on
Visual7W.

The poor performance of the~A+I transfer-learning experiment suggests that
there is a substantial difference in the answer distribution between both
datasets, especially since both use images from~\cite{lin14}.
Transferring the full model from VQA to Visual7W works
surprisingly well: we achieve $53.8\%$ accuracy, which is less than~$2\%$ worse
than LSTM-Att~\cite{zhu15}, even though the model never learns from Visual7W training data.
If we finetune the transferred model on the Visual7W dataset, it actually
outperforms a model trained from scratch on that same dataset, obtaining an
accuracy of~$\textbf{68.5\%}$. This additional boost likely stems
from the model adjusting to the biases in the Visual7W dataset.

\section{Error Analysis}
\label{Error Analysis}
\begin{figure}[h!]
\scriptsize
\centering
\kern-10pt
\begin{tabular}{m{0.22\textwidth}cm{0.22\textwidth}cm{0.22\textwidth}cm{0.22\textwidth}}
\includegraphics[width=0.22\textwidth,height=.18\textwidth]{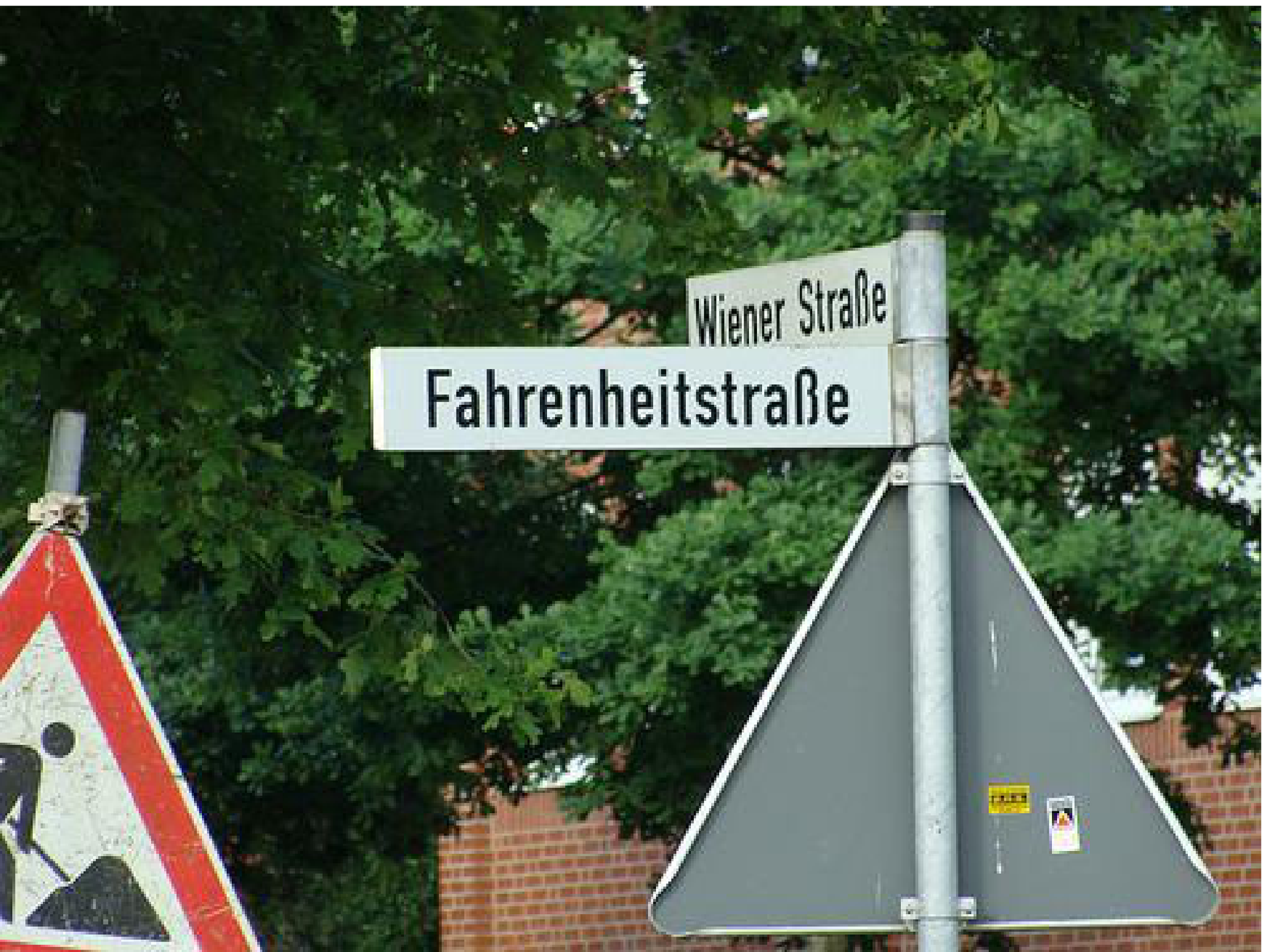} 
&~~~&
\includegraphics[width=0.22\textwidth,height=.18\textwidth]{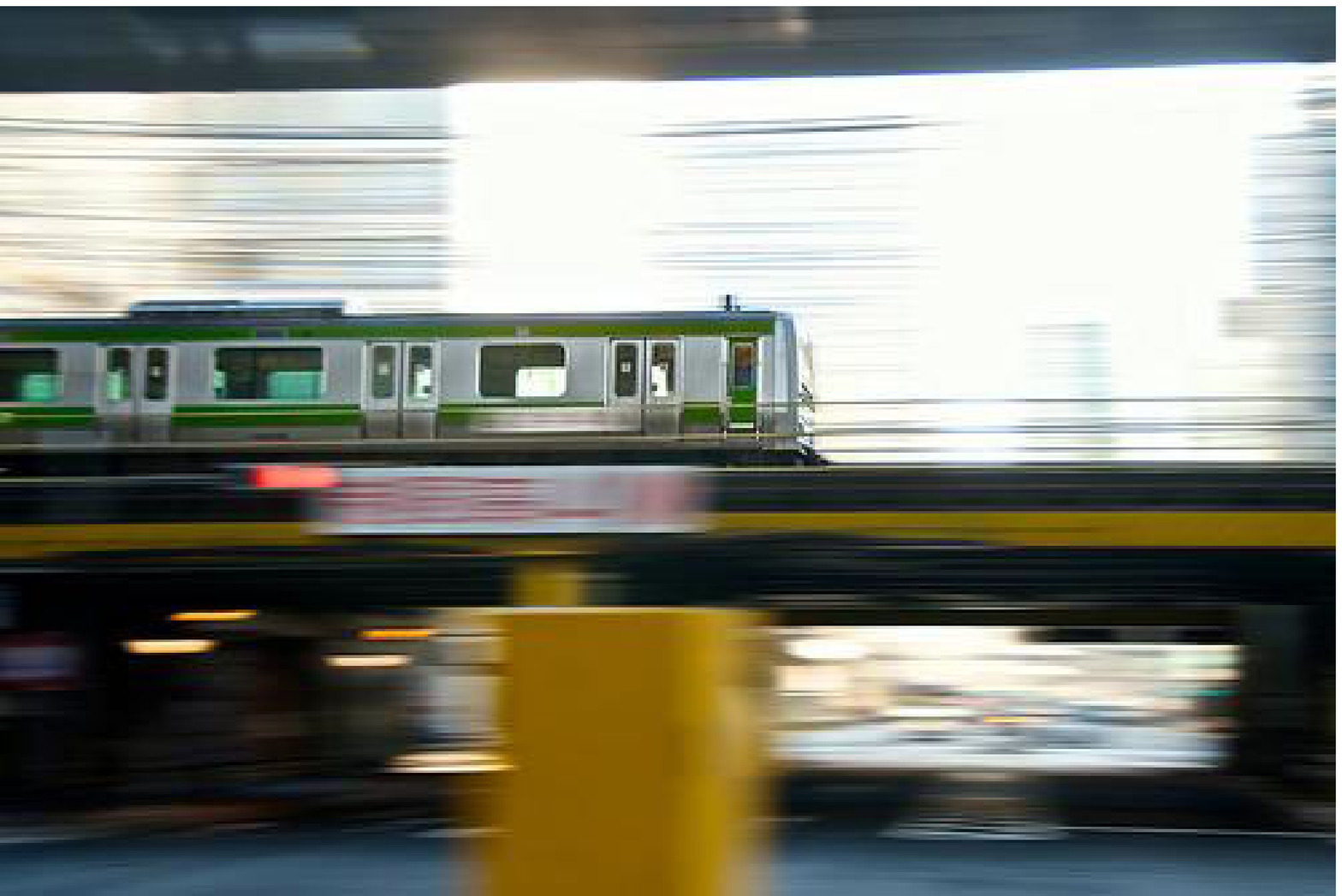} 
&~~~&
\includegraphics[width=0.22\textwidth,height=.18\textwidth]{image12.eps} 
&~~~&
\includegraphics[width=0.22\textwidth,height=.18\textwidth]{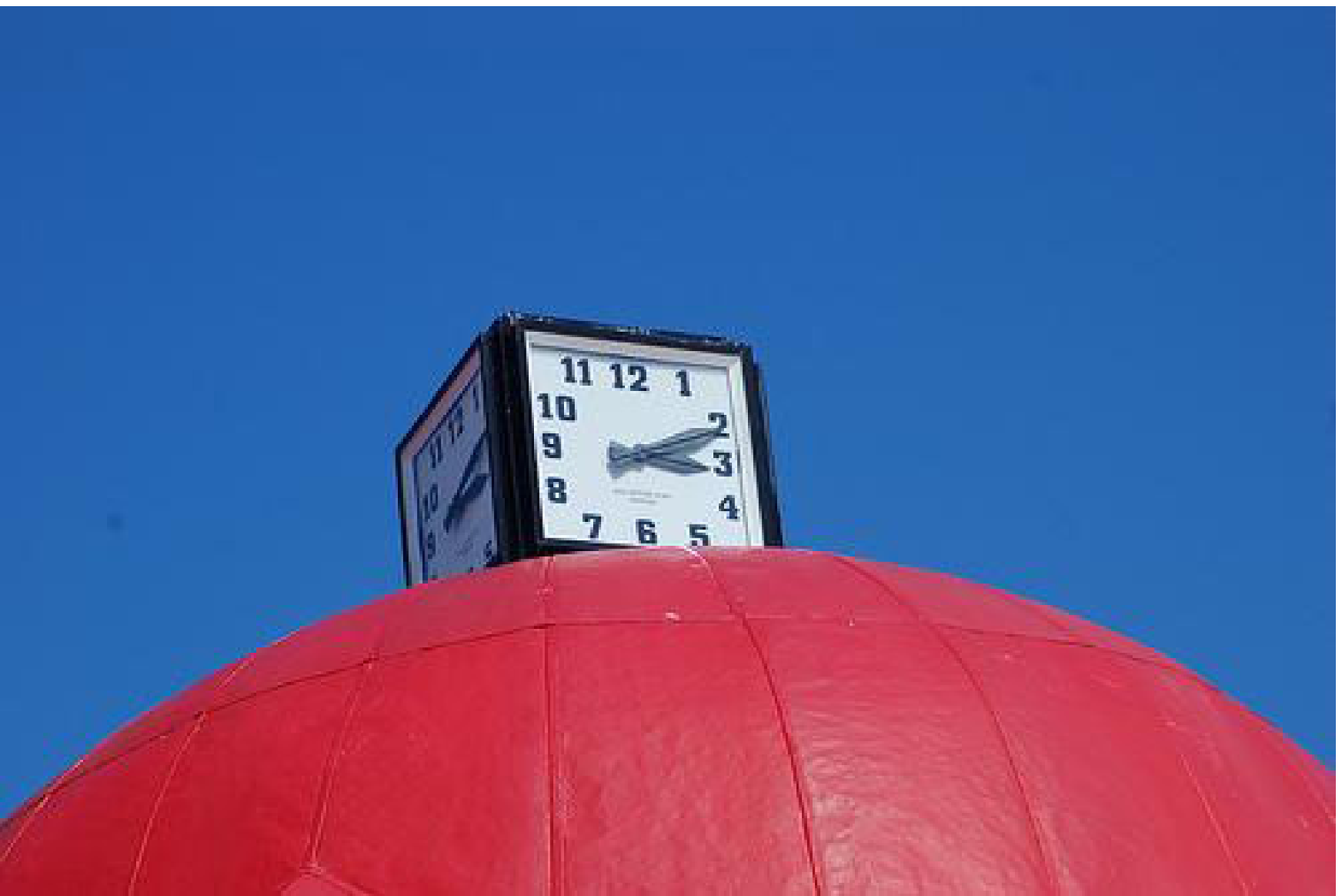} 
\\
What is the color of the tree leaves? && What is the color of the train?
&&
What shape is this sign? && What shape is the clock?
\\
{\color{ForestGreen}-Green.} && {\color{ForestGreen}-Green.} 
&&
{\color{ForestGreen}-Octagon.} && {\color{ForestGreen}-Cube.} 
\\
-Brown. && {\color{red}-Yellow.}
&&
-Oval. && {\color{red}-Circle.}
\\
-Orange. && -Black.
&&
-Hexagon. &&  -Oval.
\\
-Red. && -Red.
&&
-Square. && -Rectangle.
\end{tabular}
\caption{Examples of good and bad predictions by our visual question answering
  model on color and shape questions. Correct answers are typeset in green;
  incorrect predictions by our model are typeset in red. See text for
    details.}\label{fig:color}
\kern-20pt
\end{figure}

\begin{figure}[h!]
\scriptsize
\centering
\begin{tabular}{m{0.22\textwidth}cm{0.22\textwidth}cm{0.22\textwidth}cm{0.22\textwidth}}
\includegraphics[width=0.22\textwidth,height=.18\textwidth]{image14.eps} 
&~~~&
\includegraphics[width=0.22\textwidth,height=.18\textwidth]{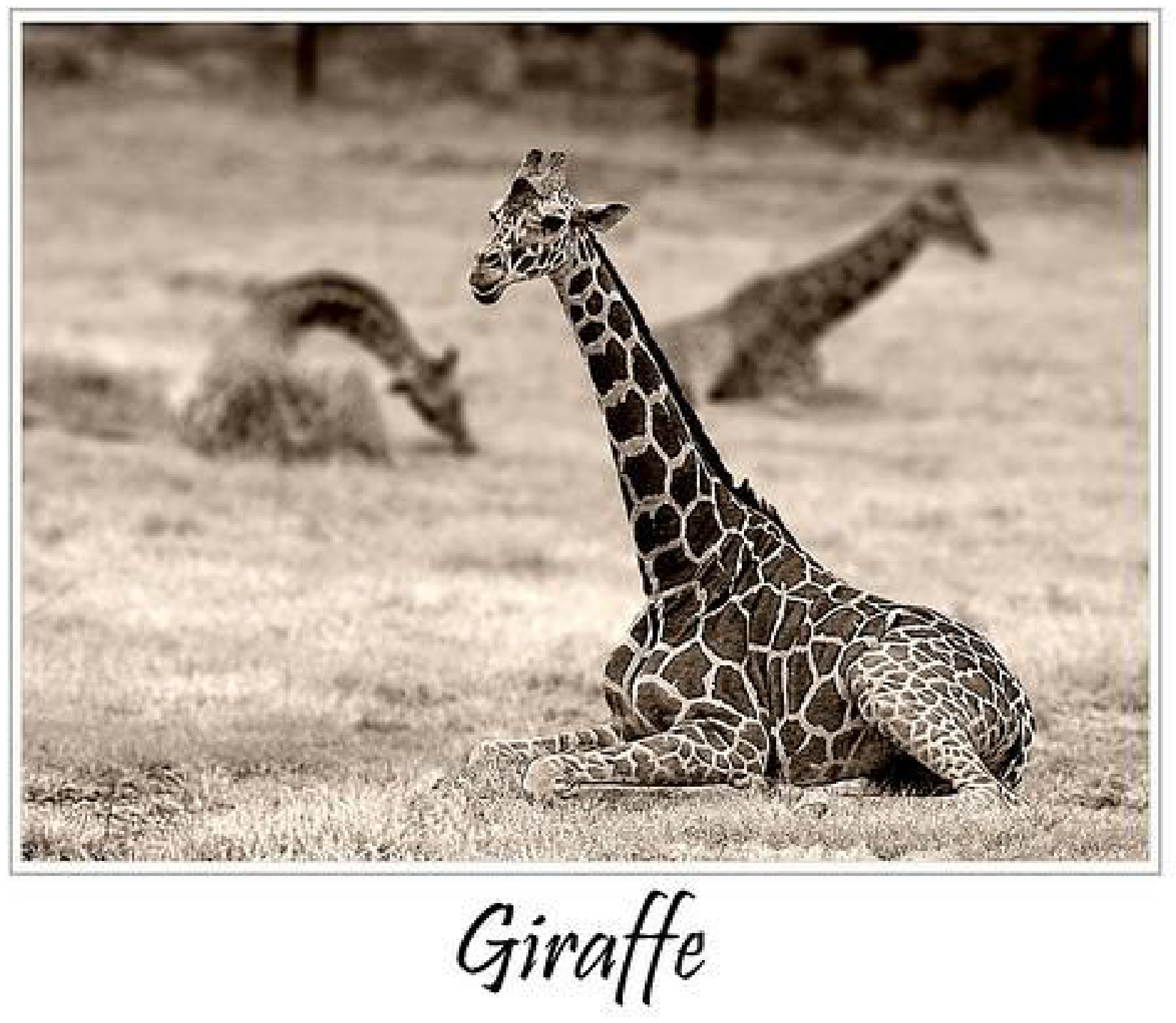} 
&~~~&
~~~~\includegraphics[width=0.16\textwidth,height=.18\textwidth]{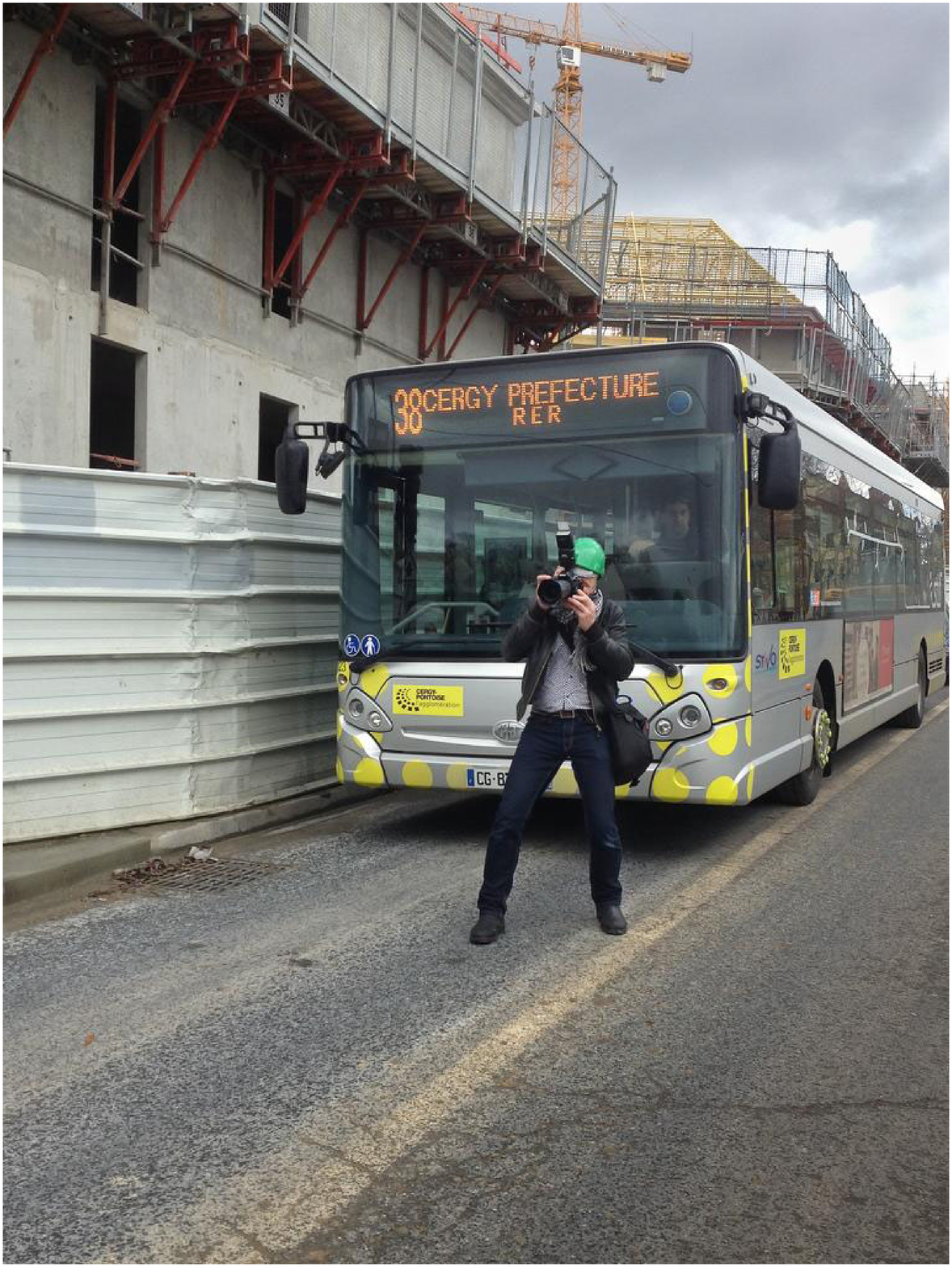} 
&~~~&
\includegraphics[width=0.22\textwidth,height=.18\textwidth]{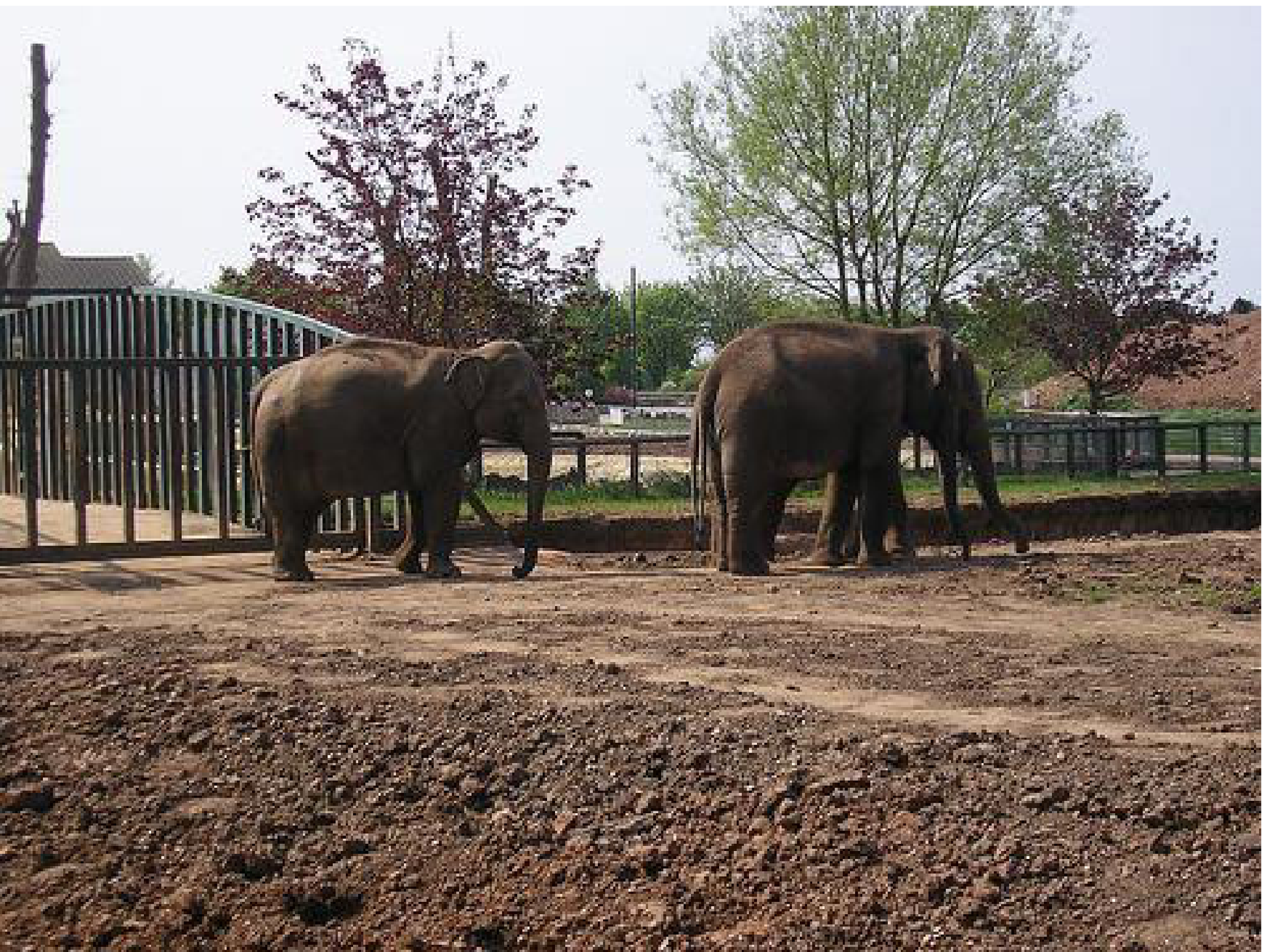} 
\\
How many clouds are in the sky? && How many giraffes sitting?
&&
What is behind the photographer? && What color leaves are on the tree behind the elephant on the left of the photo?
\\
{\color{ForestGreen}-None.} && {\color{ForestGreen}-Three.} 
&&
{\color{ForestGreen}-A bus.} && {\color{ForestGreen}-Red.} 
\\
-Three. && {\color{red}-One.}
&&
-A dump truck. && -Orange.
\\
-Five. && -Two.
&&
-A duck. && {\color{red}-Green.}
\\
-Seven. && -Four.
&&
-A plate of food. && -Brown.
\end{tabular}
\caption{Examples of good and bad predictions by our visual question answering
  model on counting and spatial reasoning. Correct answers are typeset in green;
  incorrect predictions by our model are typeset in red. See text for
    details.}\label{fig:counting}
\kern-20pt
\end{figure}

\begin{figure}[ht!]
	\scriptsize
	\centering
	\kern-10pt
	\begin{tabular}{m{0.22\textwidth}cm{0.22\textwidth}cm{0.22\textwidth}cm{0.25\textwidth}}
		\includegraphics[width=0.22\textwidth,height=.18\textwidth]{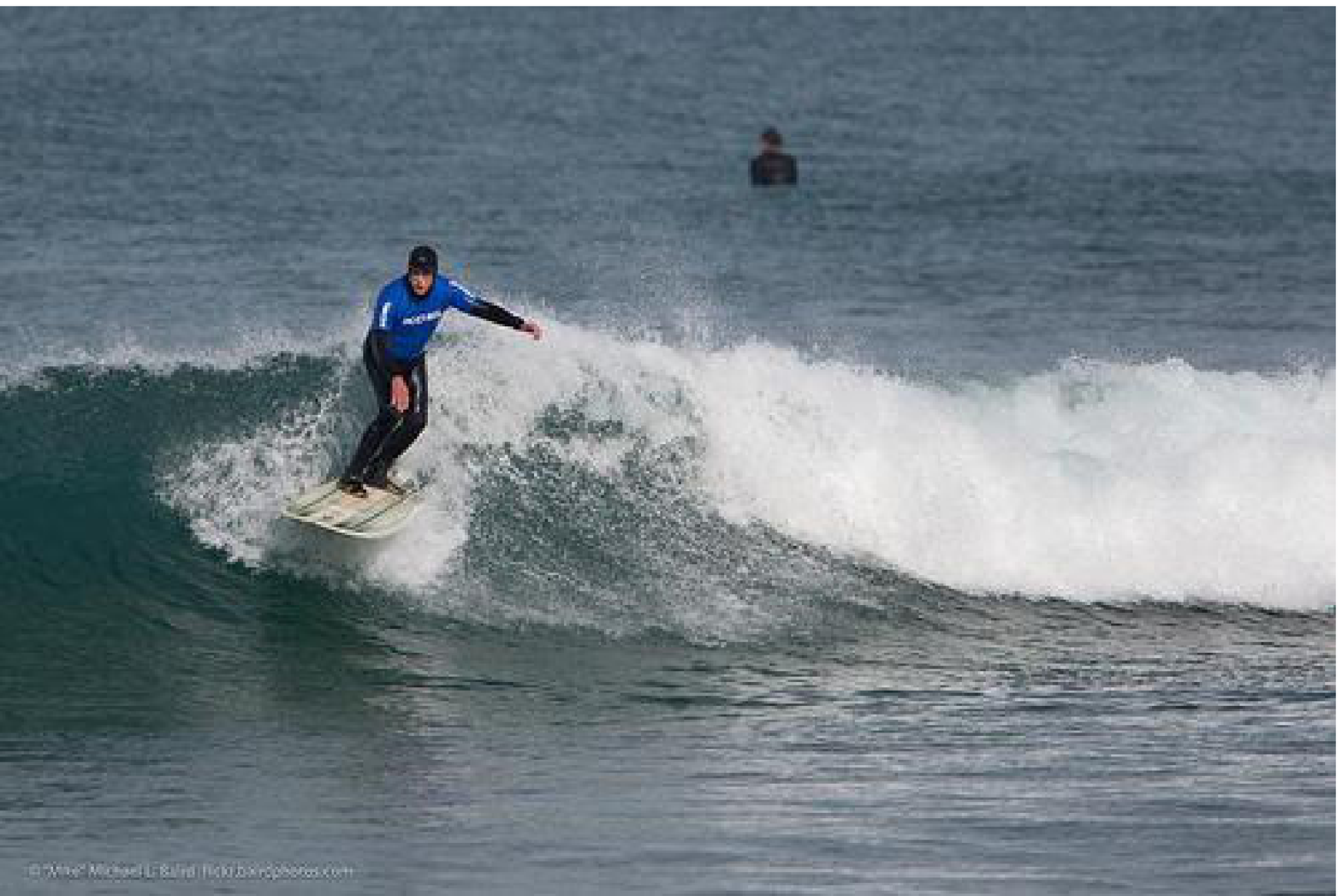} 
		&~~~& 
		~~~~\includegraphics[width=0.16\textwidth,height=.18\textwidth]{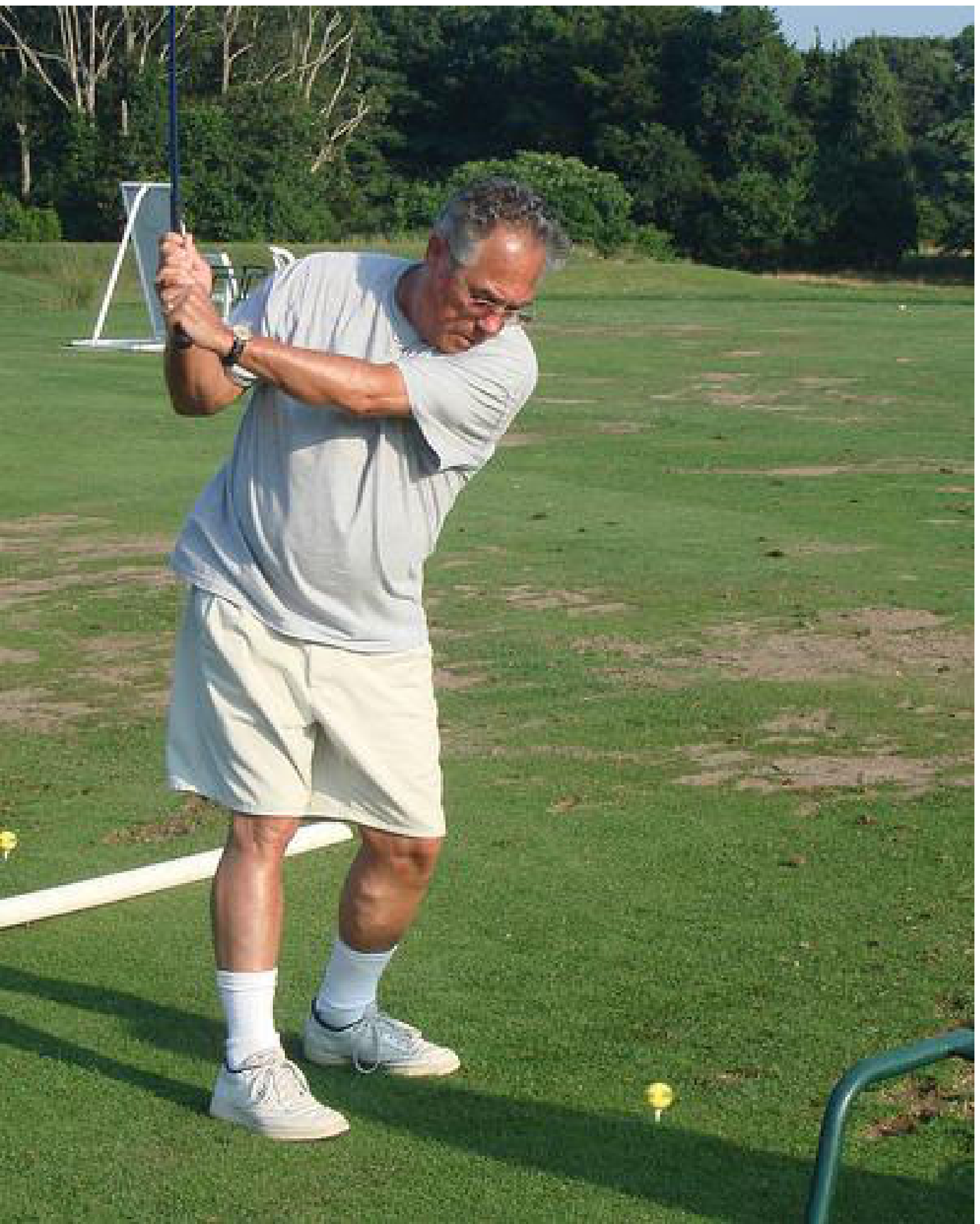} 
		&~~~&
		\includegraphics[width=0.22\textwidth,height=.18\textwidth]{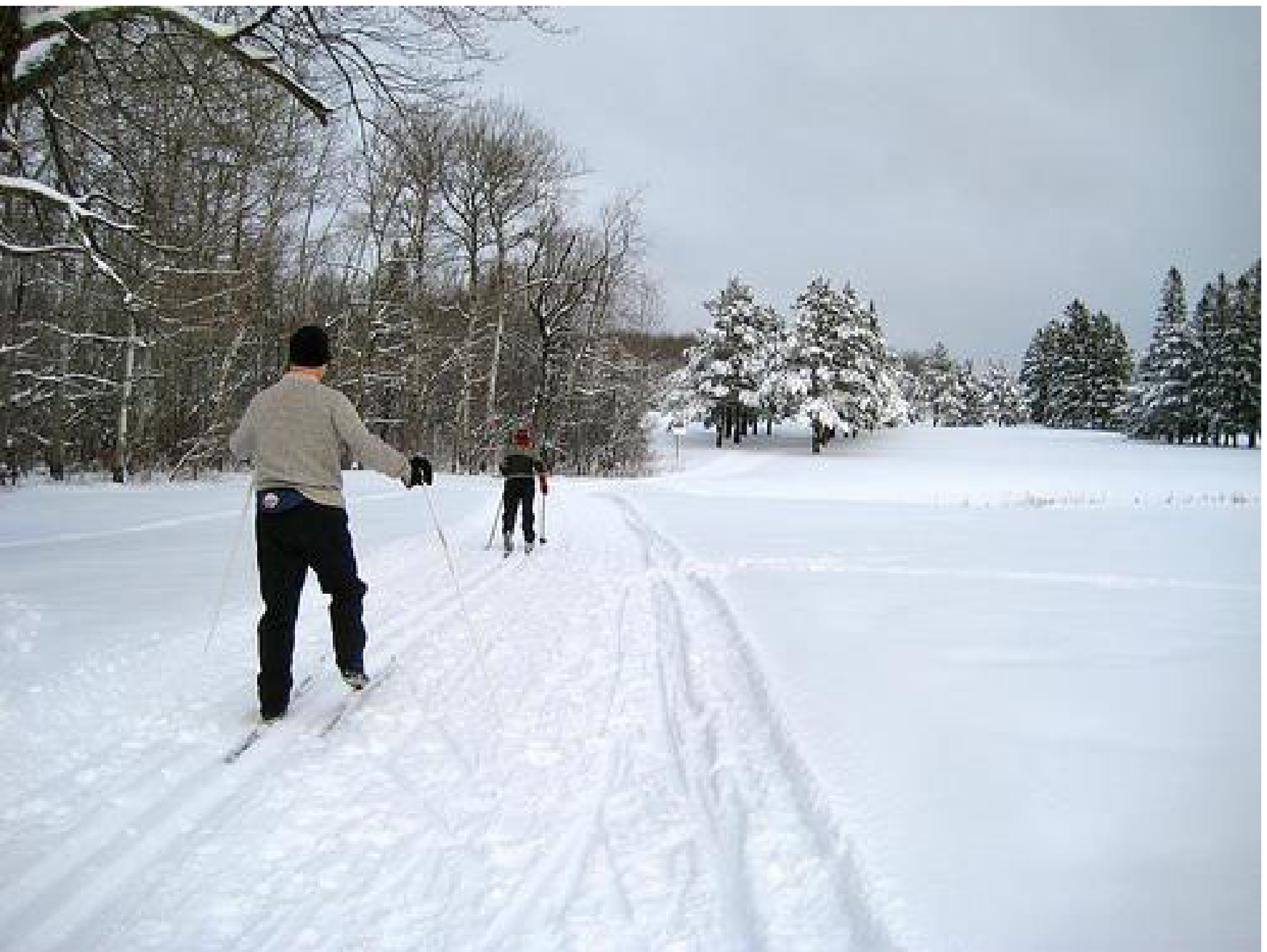} 
		&~~~&
		~~~~\includegraphics[width=0.16\textwidth,height=.18\textwidth]{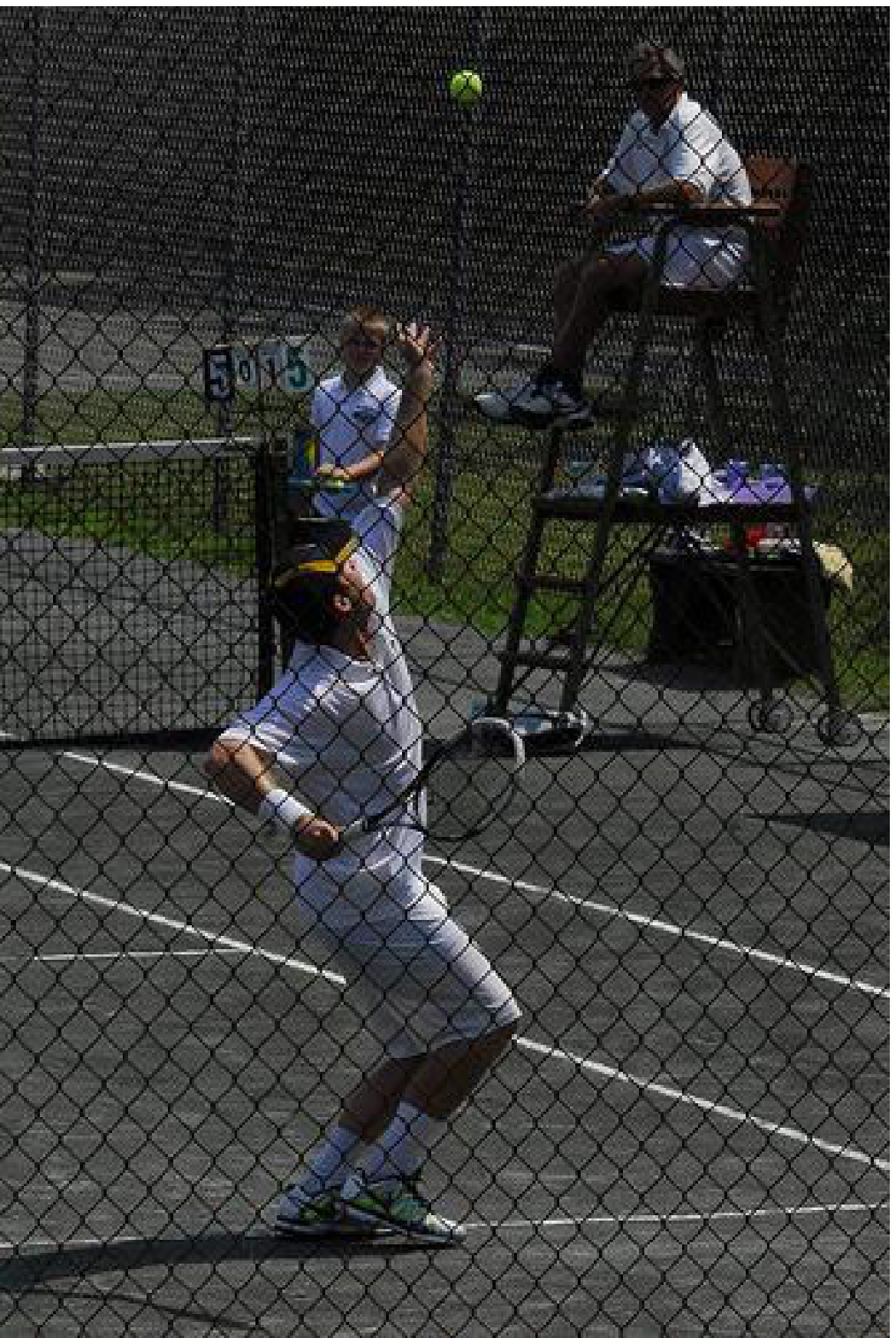} 
		\\
		What is the man doing? && What is the man doing?
		&&
		Why is the ground white? && Why is his arm up?
		\\
		{\color{ForestGreen}-Surfing.} && {\color{ForestGreen}-Golfing.} 
		&&
		{\color{ForestGreen}-Snow.} && {\color{ForestGreen}-To serve the tennis ball.} 
		\\
		-Singing. && {\color{red}-Playing tennis.}
		&&
		-Sand. && {\color{red}-About to hit the ball.}
		\\
		-Working. && -Walking.
		&&
		-Stones. && -Reaching for the ball.
		\\
		-Playing. && -Biking.
		&&
		-Concrete. && -Swinging his racket.
	\end{tabular}
	\caption{Examples of good and bad predictions by our visual question answering
		model on action and causality. Correct answers are typeset in green;
		incorrect predictions by our model are typeset in red. See text for
		details.}\label{fig:action}
	\kern-20pt
\end{figure}

To better understand the shortcomings and limitations of our models, we
performed an error analysis of the best model we obtained in
Section~\ref{Experiments} on six types of questions, which
are illustrated in Figure~\ref{fig:color}--\ref{fig:action}.


\kern3mm
\noindent\textbf{Colors and Shapes.} Approximately~$5,000$ questions in the
Visual7W test set are about colors and approximately~$200$ questions are about
shapes.  While colors and shapes are fairly simple visual features, our models
only achieve around~$57\%$ accuracy on these types of questions. For
reference, our~(A+Q) baseline already achieves~$52\%$ in accuracy.  This
means that our models primarily learn the bias in the dataset.  For example,
for shape, it predicts either~``circle",~``round", or~``octagon" when
the question is about a~``sign". 
For color questions, even though the performances are similar, it appears that
the image-based models are able to capture additional information.
For example, Figure~\ref{fig:color} shows that
the model tends to predict the most salient color, but fails to capture color
coming from small objects, which constitute a substantial number of questions in the
Visual7W dataset. This result highlights the limits of using global image
features in visual question answering.

\kern3mm
\noindent\textbf{Counting.} There are approximately~$5,000$
questions in the Visual7W test set that involve counting the number of objects
in the image~(``how many ...?'').  On this type of questions, our model
achieves an accuracy of~$56\%$. This accuracy is hardly better than that the
$55\%$ achieved by the~(Q+A) baseline.  Again, this implies that our model does
not really extract information from the image that can be used for counting. In
particular, our model has a strong preference for answers such as:~``none",~``one", or~``two". 

\kern3mm
\noindent\textbf{Spatial Reasoning.} We refer to any question that refers to a
relative position~(``left'',~``right'',~``behind'', \emph{etc.}) as questions
about~``spatial reasoning''. There are approximately~$1,500$ such questions in
the Visual7W test set. On questions requiring spatial reasoning, our models
achieve an accuracy of approximately~$55\%$, whereas a purely text-based model
achieves an accuracy~$50\%$.  This suggests that our models, indeed, extract
some information from the images that can be used to make inferences about
spatial relations. 

\kern3mm
\noindent\textbf{Actions.} We refer to any question that asks what an entity
is~``doing" as an~``action'' question. There are approximately~$1,200$
such questions in the Visual7W test set. Our models achieve an accuracy of roughly
$77\%$ on action questions. By contrast, the A+Q model achieves an
accuracy of ~$63\%$, while the A+I model achieves $75\%$. This result suggests that our model does learn to
exploit image features in recognizing actions, corroborating previous studies that show image features transfer well to
simple action-recognition tasks~\cite{joulin15,razavian14}.
\kern20pt

\kern3mm
\noindent\textbf{Causality.} ``Why'' questions test the model's ability
to capture a weak form of causality. There are around~$2,600$
of them.  Our model has an accuracy of~$68\%$ on such questions, but a simple text-based model already obtains~$64\%$. This means that most~``why'' questions can be answered by
looking at the text. This is unsurprising, as many of these questions 
refer to common sense that is encoded in the text.
For example, in Figure~\ref{fig:action}, one hardly needs the image to
correctly predict that the ground is ``white'' because of ``snow'' instead of ``sand''.

\section{Discussion and Future Work}
\label{Discussion and Future Work}
This paper presented a simple alternative model for visual question answering multiple choice, explored variants of this model, and experimented with transfer between VQA datasets. Our study produced stronger baseline systems than those presented in prior studies. In particular, our results demonstrate that featurizing the answers and training a binary classifier to predict correctness of an image-question-answer triplet leads to substantial performance improvements over the current state-of-the-art on the Visual7W Telling task: our best model obtains an accuracy of $67.1\%$ when trained from scratch, and $68.5\%$ when transferred from VQA and finetuned on the Visual7W. On the VQA Real Multiple Choice task, our model outperforms models that use LSTMs and attention mechanisms, and is close to the state-of-the-art despite being very simple.

Our error analysis demonstrates that future work in visual question answering should focus on grounding the visual entities that are present in the images, as the ``difficult'' questions in the Visual7W dataset cannot be answered without such grounding. Whilst global image features certainly help in visual question answering, they do not provide sufficient grounding of concepts of interest. More precise grounding of visual entities, as well as reasoning about the relations between these entities, is likely to be essential in making further progress. 

Furthermore, in order to accurately evaluate future models, we need to understand the biases in VQA datasets. Many of the complex methods in prior work perform worse than the simple model presented in this paper. We hypothesize that one of two things (or both) may explain these results: (1) it may be that, currently, the best-performing models are those that can exploit biases in VQA datasets the best, \emph{i.e.}, models that ``cheat'' the best; (2) it may be that current, early VQA models are unsuitable for the difficult task of visual question answering, as a result of which all of them hit roughly the same ceiling in experiments and evaluations. In some of our experiments, we have seen that a model that appears qualitatively better may perform worse quantitatively, because it captures dataset biases less well. To address such issues, it may be necessary to consider alternative evaluation criterions that are less sensitive to dataset bias.

Finally, the results of our transfer-learning experiments suggest that exploring the ability of VQA systems to generalize across datasets may be an interesting alternative way to evaluate such systems, the biases they learn, and the underlying biases of datasets on which they are trained.

\bibliographystyle{splncs}
\bibliography{egbib}
\end{document}